\def\year{2018}\relax
\documentclass[letterpaper]{article} 
\usepackage{aaai18}  
\usepackage{times}  
\usepackage{helvet}  
\usepackage{courier}  
\usepackage{url}  
\usepackage{graphicx}  

\nocopyright 

\usepackage{multibib}
\usepackage[russian,hindi,french,spanish,italian,english]{babel}
\usepackage[utf8]{inputenc}
\newcites{languageresource}{Language Resources}
\usepackage{graphicx}
\usepackage{tabularx}
\usepackage{soul}
\usepackage{latexsym}
\usepackage{enumerate}
\usepackage[inline]{enumitem}
\usepackage{tabularx}
\usepackage{booktabs} 
\usepackage{subcaption}
\usepackage{multirow}
\usepackage[colorinlistoftodos,prependcaption,textsize=footnotesize,color=green]{todonotes}
\usepackage{scrextend}
\usepackage{microtype}
\newcommand{\citet}[1]
{\citeauthor{#1}~\shortcite{#1}}
\newcommand{\citep}{\cite}
\newcommand{\citealp}[1]
{\citeauthor{#1}~\citeyear{#1}}
\begin{document}
%
\title{280 Birds with One Stone: Inducing Multilingual Taxonomies from Wikipedia Using Character-level Classification}
\author{Amit Gupta and R\'emi Lebret and Hamza Harkous and Karl Aberer\\
\'Ecole Polytechnique F\'ed\'erale de Lausanne\\
 Route Cantonale, 1015 Lausanne \\
}
\maketitle
\newcommand\BibTeX{B{\sc ib}\TeX}
\newcommand{\isa}{\textit{is-a}}
\newcommand{\notisa}{\textit{not-is-a}}
\newcommand{\mwb}{MultiWiBi}
\newcommand{\mta}{MENTA}

\begin{abstract}
We propose a novel fully-automated approach towards inducing multilingual taxonomies from Wikipedia. Given an English taxonomy, our approach first leverages the interlanguage links of Wikipedia to automatically construct training datasets for the \isa{} relation in the target language. Character-level classifiers are trained on the constructed datasets, and used in an optimal path discovery framework to induce high-precision, high-coverage taxonomies in other languages. Through experiments, we demonstrate that our approach significantly outperforms the state-of-the-art, heuristics-heavy approaches for six languages. As a consequence of our work, we release presumably the largest and the most accurate multilingual taxonomic resource spanning over 280 languages.
\end{abstract}

\section{Introduction}
\paragraph{Motivation.}Machine-readable semantic knowledge in the form of taxonomies (i.e., a collection of \isa{}\footnote{We use the terms \isa{} and \textit{hypernym} interchangeably.} edges) has proved to be beneficial in an array of Natural Language Processing (NLP) tasks, including inference, textual entailment, question answering, and information extraction~\cite{biemann2005ontology}. This has led to multiple large-scale manual efforts towards taxonomy induction such as WordNet~\cite{miller1995wordnet}. However, manual construction of taxonomies is time-intensive, usually requiring massive annotation efforts. Furthermore, the resulting taxonomies suffer from low coverage and are unavailable for specific domains or languages. Therefore, in the recent years, there has been substantial interest in inducing taxonomies automatically, either from unstructured text~\cite{velardi2013ontolearn},
or from semi-structured collaborative content such as Wikipedia~\cite{hovy2013collaboratively}.

Wikipedia, the largest publicly-available source of multilingual, semi-structured content~\cite{remy2002}, has served as a key resource for automated knowledge acquisition. One of its core components is the Wikipedia Category Network (hereafter referred to as \textbf{WCN}), a semantic network which links Wikipedia entities\footnote{We use Wikipedia \textit{page} and \textit{entity} interchangeably.}, such as \textit{Johnny Depp}, with inter-connected categories of different granularity (e.g., \textit{American actors}, \textit{Film actors}, \textit{Hollywood}). The semi-structured nature of WCN has enabled the acquisition of large-scale taxonomies using lightweight rule-based approaches~\cite{hovy2013collaboratively}, thus leading to a consistent body of research in this direction. 

The first line of work on taxonomy induction from Wikipedia mainly focuses on the English language. This includes WikiTaxonomy~\cite{ponzetto2008wikitaxonomy}, WikiNet~\cite{nastase2010wikinet}, YAGO~\cite{suchanek2007yago,hoffart2013yago2}, DBPedia~\cite{auer2007dbpedia}, and Heads Taxonomy~\cite{guptarevisiting}. 

The second line of work aims to exploit the multilingual nature of Wikipedia. MENTA~\cite{de2010menta}, one of the largest multilingual lexical knowledge bases, is constructed by linking WordNet and Wikipedias of different languages into a single taxonomy. Similarly, YAGO3~\cite{mahdisoltani2014yago3} extends YAGO by linking Wikipedia entities in multiple languages with WordNet. The most recent approach to multilingual taxonomy induction from Wikipedia is the Multilingual Wikipedia Bitaxonomy Project or \mwb{}~\cite{flati2016multiwibi}. \mwb{} first induces taxonomies for English, which are further projected to other languages using a set of complex heuristics that exploit the interlanguage links of Wikipedia. Unlike MENTA and YAGO3, \mwb{} is self-contained in Wikipedia, i.e., it does not require labeled training examples or external resources such as WordNet or Wikitionary. While \mwb{} is shown to outperform MENTA and YAGO3 considerably, it still achieves low precision for non-English pages that do not have an interlanguage link to English (e.g., 59\% for Italian).


\vspace{-0.2cm}
\paragraph{Contributions.} 
In this paper, we propose a novel approach towards inducing multilingual taxonomies from Wikipedia. Our approach is fully-automated and language-independent. It provides a significant advancement over state of the art in multilingual taxonomy induction from Wikipedia because of the following reasons:

\begin{itemize}
\item 

Most previous approaches such as MENTA or \mwb{} rely on a set of complex heuristics that utilize custom hand-crafted features. In contrast, our approach employs text classifiers in an optimal path search framework to induce taxonomies from the WCN. The training set for text classifiers is automatically constructed using the Wikipedia interlanguage links. As a result, our approach is simpler, more principled and easily replicable.
\item Our approach significantly outperforms the state-of-the-art
approaches across multiple languages in both (1) standard edge-based precision/recall measures and (2) path-quality measures. Furthermore, our taxonomies have significantly higher branching factor than the state-of-the-art taxonomies without incurring any loss of precision.
\item As a consequence of our work, we release presumably the largest and the most accurate multilingual taxonomic resource spanning over 280 languages. We also release edge-based gold standards for three different languages (i.e., French, Italian, Spanish) and annotated path datasets for six different languages (i.e., French, Italian, Spanish, Chinese, Hindi, Arabic) for further comparisons and benchmarking purposes.

\end{itemize}

\section{Taxonomy Induction}
\label{sec:taxo}
\paragraph{Background.} We start by providing a description of the various components of Wikipedia, which will aid us in presenting the rest of this paper:
\begin{itemize}[leftmargin=0.2cm,itemsep=0.9pt,topsep=0.8pt]
\setlength{\itemindent}{1.4em}
\item A Wikipedia \textbf{page} describes a single entity or a concept. Examples of pages include \textit{Johnny Depp}, \textit{Person}, or \textit{Country}. Currently, Wikipedia consists of more than 44 million pages spanning across more than 280 different languages~\cite{wikixxx}.  

\item A Wikipedia \textbf{category} groups related pages and other categories into broader categories. For example, the category \textit{American actors} groups pages for American actors, such as \textit{Johnny Depp}, as well as other categories, such as \textit{American child actors}. 
The directed graph formed by pages and categories as nodes, and the groupings as edges is known as the \textbf{Wikpedia Category Network (WCN)}. A different WCN exists for each of the 280 languages of Wikipedia. WCN edges tend to be noisy, and are usually a mix of \isa{} (e.g., \textit{Johnny Depp}$\rightarrow$\textit{American actors}) and \notisa{} edges (e.g., \textit{Johnny Depp}$\leadsto$\textit{Hollywood}). 

\item An \textbf{Interlanguage link} connects a page (or a category) with their equivalent page (or category) across different languages. For example, the English page for \textit{Johnny Depp} is linked to its equivalent versions in 49 different languages including French (\textit{Johnny Depp}) and Russian (\foreignlanguage{russian}{Депп, Джонни}). Two nodes linked by an interlanguage link are hereafter referred to as \textit{\textbf{equivalent}} to each other. 
\end{itemize}

\paragraph{Algorithm.} We now describe our approach for inducing multilingual taxonomies from the WCN. Given (1) a unified taxonomy of pages and categories in English (we use Heads Taxonomy publicly released by~\citet{guptarevisiting}\footnote{We note that our method is independent of the English taxonomy induction method.}), (2) the interlanguage links, and (3) a target language, our approach aims to induce a unified taxonomy of pages and categories for the target language. Our approach runs in three phases: 
\begin{enumerate}[label=\roman*),leftmargin=0.2cm,itemsep=1pt,topsep=0.8pt]
\setlength{\itemindent}{1.4em}
\item \textbf{Projection phase}: create a high-precision, low-coverage taxonomy for the target language by projecting \isa{} edges from the given English taxonomy using the interlanguage links.
\item \textbf{Training phase}: leverage the high-precision taxonomy to train classifiers for classifying edges into \isa{} or \notisa{} in the target language.
\item \textbf{Induction Phase}: induce the final high-precision, high-coverage taxonomy by running optimal path search over the target WCN with edge weights computed using the trained classifiers.

\end{enumerate}

\subsection{Projection Phase} 
\label{sec:proj}
Let $T_e$ be the given English taxonomy. Let $G_f$ be the WCN and $T_f$ be the output taxonomy (initially empty) for the target language $f$ (such as French). 
For a node $n_f\in G_f$ with the English equivalent $n_e$, for which no hypernym exists yet in $T_f$, we perform the following steps: 

\begin{enumerate}[label=\roman*),leftmargin=0.2cm,itemsep=1pt,topsep=0.5pt]
\setlength{\itemindent}{1.4em}
\item
Collect the set $A_e$ of all ancestor nodes of $n_e$ in $T_e$ up to a fixed height $k_1$\footnote{In our experiments, $k_1=14$ sufficed as Heads taxonomy had a maximum height of $14$ and no cycles.}. 
\item
Fetch the set $A_f$ of equivalents for nodes in $A_e$ in the target language $f$. 
\item
Find the shortest path between $n_f$ and any node in $A_f$ up to a fixed height $k_2$\footnote{$k_2$ is set to 3 to maintain high precision.}; 
\item
Add all the edges in the shortest path to the output taxonomy $T_f$. 
\end{enumerate}

\begin{figure}[t]
\centering
\includegraphics[width=3.3in]{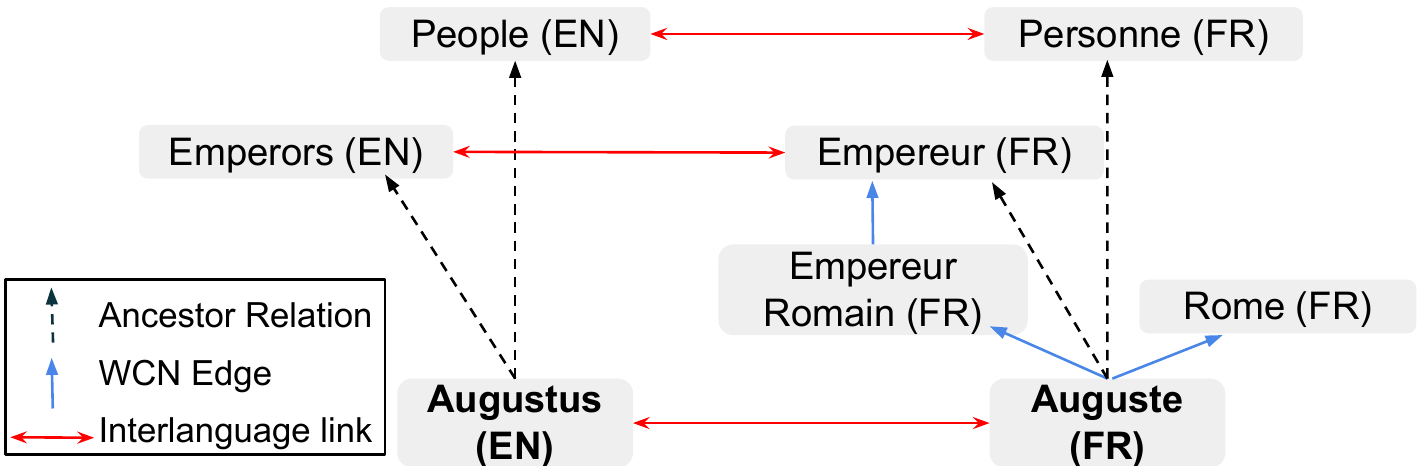}
\caption{Example of projection phase.}
\label{fig:proj_example}
\end{figure}

If no English equivalent $n_e$ exists, the node $n_f$ is ignored. Figure~\ref{fig:proj_example} shows an example of the projection phase with French as the target language. For French node \textit{Auguste}, its English equivalent (i.e., \textit{Augustus}) is fetched via the interlanguage link. The ancestors of \textit{Augustus} in English taxonomy (i.e., \textit{Emperors}, \textit{People}) are collected, and mapped to their French equivalents (i.e., \textit{Empereur}, \textit{Personne}). Finally, the WCN edges in the shortest path from \textit{Auguste} to \textit{Empereur} (i.e., \textit{Auguste}$\rightarrow$\textit{Empereur Romain}, \textit{Empereur Romain}$\rightarrow$\textit{Empereur}) are added to the output French taxonomy.

\subsection{Training Phase} 
\label{sec:trainp}

Up till now, we constructed an initial taxonomy for the target language by simply projecting the English taxonomy using the interlanguage links. However, the resulting taxonomy suffers from low coverage, because nodes that do not have an English equivalent are ignored. For example, only 44.8\% of the entities and 40.5\% of the categories from the French WCN have a hypernym in the projected taxonomy. 

Therefore, to increase coverage, we train two different binary classifiers for classifying remaining target WCN edges into \isa{} (positive) or \notisa{} (negative). The first classifier is for Entity$\rightarrow$Category edges and the other for Category$\rightarrow$Category edges\footnote{Entity$\rightarrow$Entity and Category$\rightarrow$Entity edges are not present in the WCN.}. We construct the training data for edge classification as follows: 
\begin{enumerate}[label=\roman*),leftmargin=0.2cm,itemsep=1pt,topsep=0.5pt]
\setlength{\itemindent}{1.4em}
\item
Assign an \isa{} label to the edges in $T_f$ (i.e., the projected target taxonomy).
\item
 Assign a \notisa{} label to all the edges in $G_f$ (i.e., the target WCN) that are not in $T_f$ but originate from a node covered in $T_f$.

\end{enumerate}

For example, in Figure~\ref{fig:proj_example}, the edge \textit{Auguste}$\rightarrow$\textit{Empereur Romain} is assigned the \isa{} label, and other WCN edges starting from \textit{Auguste} (e.g., \textit{Auguste}$\rightarrow$\textit{Rome}) are assigned the \notisa{} label.

\paragraph{Classifiers.}To classify edges into \isa{} or \notisa{}, we experiment with the following classifiers:

\begin{enumerate}[label=\roman*),leftmargin=0.2cm,itemsep=0.9pt,topsep=0.8pt]
\setlength{\itemindent}{1.4em}
\item
\textbf{Bag-of-words TFIDF}: Given edge $A$$\rightarrow$$B$, concatenate the features vectors for $A$ and $B$ computed using TFIDF over bag of words, and train a linear Support Vector Machine over the concatenated features. This method is hereafter referred to as \textbf{Word TFIDF}.
\item
\textbf{Bag-of-character-ngrams TFIDF:} Same as Word TFIDF, except TFIDF is computed over bag of character $n$-grams\footnote{\label{note1}$n$-values=\{2,3,4,5,6\} worked best in our experiments.} (hereafter referred to as \textbf{Char TFIDF}).
\item \textbf{fastText:} A simple yet efficient baseline for text classification based on a linear model with a rank constraint and a fast loss approximation. Experiments show that fastText typically produces results on par with sophisticated deep learning classifiers~\cite{joulin2016bag}.
\item \textbf{Convolutional Neural Network (CNN):} We use a single-layer CNN model trained on top of word vectors as proposed by~\citet{Kim14}. We also experiment with a character version of this model, in which instead of words, vectors are computed using characters and fed into the CNN. These models are referred to as \textbf{Word CNN} and \textbf{Char CNN} respectively. Finally, we experiment with a two-layer version of the character-level CNN proposed by~\cite{ZhangZL15}, hereafter referred to as \textbf{Char CNN-2l}. 
\item \textbf{Long Short Term Memory Network (LSTM):} We experiment with both word-level and character-level versions of LSTM~\cite{HochreiterS97}. These models are hereafter referred to as \textbf{Word LSTM} and \textbf{Char LSTM} respectively. 

\end{enumerate}

\subsection{Induction Phase}
\label{sec:indp}
In the last step of our approach, we discover taxonomic edges for nodes not yet covered in the projected taxonomy ($T_f$). To this end, we first set the weights of Entity$\rightarrow$Category and Category$\rightarrow$Category edges in the target WCN as the probability of being \isa{} (computed using the corresponding classifiers). Further, for each node $n_f$ without a hypernym in $T_f$, we find the top $k$ paths\footnote{$k$ is set to 1 unless specified otherwise.} with the highest probabilities originating from $n_f$ to any node in $T_f$, where the probability of a path is defined as the product of probabilities of individual edges\footnote{If multiple paths with the same probabilities are found, the shortest paths are chosen.}. The individual edges of the most probable paths are added to the final taxonomy.

\label{sec:bfs}

\section{Evaluation}
\label{sec:eval}
In this section, we compare our approach against the state of the art using two different evaluation methods. In Section~\ref{sec:edgeeval}, we compute standard edge-level precision, recall, and coverage measures against a gold standard for three languages. In section~\ref{sec:patheval}, we perform a comprehensive path-level comparative evaluation across six languages. We compare our approach against \mwb{} due to the following reasons:

\begin{itemize}[leftmargin=0.2cm,topsep=0.5pt]
\setlength{\itemindent}{1.4em}
\item Only MENTA, \mwb{}, and our taxonomies are constructed in a fully language-independent fashion; hence, they are available for all 280 Wikipedia languages.
\item Unlike YAGO3, MENTA and most other approaches, \mwb{} and ours are self-contained in Wikipedia. They do not require manually labeled training examples or external resources, such as WordNet or Wikitionary. 
\item \mwb{} has been shown to outperform all other previous approaches including YAGO3 and MENTA~\cite{flati2016multiwibi}.
\end{itemize}

\subsection{Edge-level Evaluation}
\label{sec:edgeeval}
\paragraph{Experimental Setup.} We faced a tough choice of selecting a Wikipedia snapshot since \mwb{}, to which we compare, is constructed using a 2012 snapshot whereas~\citet{guptarevisiting}, on which we build, uses a 2015 snapshot. Additionally, the code, executable, and gold standards used by \mwb{} were not available upon request. Therefore, to advance the field and produce a more recent resource, we decided to use a 2015 snapshot of Wikipedia, especially given that~\citet{guptarevisiting} point out that there is no evidence that taxonomy induction is easier on recent editions of Wikipedia.

We create gold standards for three languages (French, Spanish and Italian) by selecting 200 entities and 200 categories randomly from the November 2015 snapshot of Wikipedia and annotating the correctness of the WCN edges originating from them\footnote{Two annotators independently annotated each edge. Inter-annotator agreement (Cohen's Kappa) varied between 0.71 to 0.93 for different datasets.}. Table~\ref{tab:sample_edges} shows a sample of annotated edges from the French gold standard. In total, 4045 edges were annotated across the three languages. 

For evaluation, we use the same metrics as \mwb{}: (1) Macro-precision ($P$) defined as the average ratio of correct hypernyms to the total number of hypernyms returned (per node), (2) Recall ($R$) defined as the ratio of nodes for which at least one correct hypernym is returned, and (3) Coverage ($C$) defined as the ratio of nodes with at least one hypernym returned irrespective of its correctness.
\paragraph{Training Details.} All neural network models are trained on Titan X (Pascal) GPU using the Adam optimizer~\cite{KingmaB14}. Grid search is performed to determine the optimal values of hyper-parameters. For CNN models, we use an embedding of 50 dimensions. The number of filters is set to 1024 for word-level models and 512 for character-level models. For Char CNN-2l model, we use the same parameters used in~\citet{ZhangZL15}. For LSTM models, we use an embedding of 128 dimensions, and 512 units in the LSTM cell. We also experimented with more complex architectures, such as stacked LSTM layers and bidirectional LSTMs. However, these architectures failed to provide any significant improvements over the simpler ones.

\begin{table}[t]
\centering
\resizebox{0.9\linewidth}{!}{
\begin{tabular}{c}
\toprule
{\textbf \isa{}} \\
\midrule
Naissance à Omsk$\rightarrow$Naissance en Russie par ville \\
Port d'Amérique du Sud$\rightarrow$Port par continent  \\
\midrule
{\textbf \notisa{}} \\
\midrule
Naissance à Omsk$\leadsto$Omsk \\
Port d'Amérique du Sud$\leadsto$Géographie de l'Amérique du Sud \\
\bottomrule
\end{tabular}
}
\caption{Examples of Annotated Edges (French).}
\label{tab:sample_edges}
\end{table}

\begin{table}[t]
\centering
\resizebox{\linewidth}{!}{
\begin{tabular}{lrccc|ccc}
\toprule
&  & \multicolumn{3}{c|}{Entity} & \multicolumn{3}{c}{Category} \\
Language & Method & P & R & C & P & R & C\\
\midrule
\multirow{7}{*}{French} 
& Original WCN & 72.0 & 100 & 100 & 78.8 & 100 & 100 \\
& MENTA & 81.4 &	48.8 & 59.8  &  82.6 &	55.0 &	65.7 \\
& MultiWiBi & 84.5 & 80.9 & 94.1 & 80.7 & 80.7 & 100 \\
& UNIFORM & 80.6 & 83.2 & 100 & 85.7 & 86.7 & 100\\
& Word TFIDF & 86.5 & 90.1 & 100 & 82.1 & 83.1 & 100\\
& Char TFIDF & \underline{\textbf{88.0}} & \underline{\textbf{91.7}} & 100 & 92.3 &  93.4 & 100\\
& fastText & 86.5 & 90.1 & 100 & 90.5 & 91.6 & 100 \\
& Word LSTM & \textbf{87.8} & \textbf{91.5} & 100 & 91.6& 92.7& 100 \\
& Char LSTM & 86.2 & 89.8 &  100 &\underline{\textbf{93.9}} & \underline{\textbf{95.1}}&  100 \\
& Word CNN & 86.3 & 90.0 & 100 & \textbf{92.8} & \textbf{93.9} & 100 \\
& Char CNN & 86.2& 89.9&  100 & \textbf{93.3} &\textbf{94.4} & 100 \\
& Char CNN-2l & \textbf{87.7} &\textbf{91.0}& 100  & 92.2 & 93.3 & 100\\ 
\midrule
\multirow{7}{*}{Italian} 
& Original WCN & 74.5 & 100 & 100 & 76.2 & 100 & 100 \\
& MENTA & 79.7 &	53.2 &	66.7 &  77.1 &	25.4 &	32.8 \\
& MultiWiBi & 80.1 & 79.4 & 96.3 & \underline{\textbf{89.7}} & \textbf{89.0} & 99.2 \\
& UNIFORM & 77.7 & 81.6 & 100 & 86.6 & 88.3 & 100\\
& Word TFIDF & \textbf{90.0} & \textbf{94.4} & 100 & 84.1 & 85.7 & 100\\
& Char TFIDF & 88.4 & 92.8 & 100 & \textbf{89.2} & \underline{\textbf{90.9}} & 100\\
& fastText & 86.8 & 91.1 & 100 & \textbf{87.3} & \textbf{89.0} & 100 \\
& Word LSTM & \textbf{90.9} & \textbf{95.4} & 100 & 83.1 & 84.8 & 100 \\
& Char LSTM  & 89.8 & 94.4  &  100 & 83.3 & 83.8 &  100 \\
& Word CNN & 89.6 & 94.3 & 100 & 83.1 & 84.8 & 100 \\
& Char CNN & \underline{\textbf{92.6}} & \underline{\textbf{97.2}} &  100 & 86.9 &  88.7& 100 \\
& Char CNN-2l & 87.7& 92.1 & 100  & 86.1 & 87.8 & 100 \\ 
\midrule
\multirow{7}{*}{Spanish} 
& Original WCN & 81.4 & 100 & 100 & 80.9 & 100 & 100 \\
& MENTA & 81.0 &	42.9 &	52.7 & 80.5 & 54.2 &	66.4 \\
& MultiWiBi & 87.0 & 82.0 & 93.7 & 84.8 & 84.4 & 100\\
& UNIFORM & 88.0 & 90.7 & 100 & 83.0 & 85.0 & 100 \\
& Word TFIDF & 89.9 & 92.7 & 100 & 78.9 & 80.8 & 100\\
& Char TFIDF & 92.5 & 95.4 & 100 & 88.3 & 90.4 & 100\\
& fastText & \textbf{93.0} & \textbf{95.9} & 100 & \textbf{88.9} & \textbf{91.0} & 100 \\
& Word LSTM & \underline{\textbf{93.4}} & \textbf{96.3}  & 100 & 88.2 & 90.3 & 100 \\
& Char LSTM  &92.3 & 95.3&  100 &88.8 & 90.3&  100 \\
& Word CNN & 92.9 & 95.8 & 100 & 87.6& 89.7& 100 \\
&Char CNN & 92.9 & 95.8&  100 & \underline{\textbf{92.9}} & \underline{\textbf{95.1}}& 100 \\
&Char CNN-2l & \textbf{93.3} & \underline{\textbf{96.3}} &100  & \textbf{89.9} & \textbf{92.1} & 100\\ 
\bottomrule
\end{tabular}
}
\caption{Edge-level precision (P), recall (R) and Coverage (C) scores for different methods. MENTA and MultiWiBi results as reported by~\citet{flati2016multiwibi}. The top 3 results are shown in bold, and the best is also underlined.} 
\label{table:edgeeval}
\end{table}

\paragraph{Results.}Table~\ref{table:edgeeval} shows the results for different methods including the state-of-the-art approaches (i.e., MENTA and \mwb{}) and multiple versions of our three-phase approach with different classifiers. It also includes two baselines, i.e., \textbf{WCN} and \textbf{UNIFORM}. The WCN baseline outputs the original WCN as the induced taxonomy without performing any kind of filtering of edges. UNIFORM is a uniformly-random baseline, in which all the edge weights are set to 1 in the induction phase (cf. Section~\ref{sec:indp}).

Table~\ref{table:edgeeval} shows that all classifiers-based models achieve significantly higher precision than UNIFORM and WCN baselines, thus showing the utility of weighing with classification probabilities in the Induction phase. Interestingly, UNIFORM achieves significantly higher precision than WCN for both entities and categories across all three languages, hence, demonstrating that optimal path search in the Induction phase also contributes towards hypernym selection.
All classifier-based approaches (except Word TFIDF) significantly outperform \mwb{} for entities across all languages as well as for French and Spanish categories. Although \mwb{} performs better for Italian categories, Char TFIDF achieves similar performance (89.2\% vs 89.7\%) \footnote{We note that entity edges are qualitatively different for \mwb{} and other methods, i.e., \mwb{} has Entity$\rightarrow$Entity edges whereas other methods have Entity$\rightarrow$Category edges. 
Given that fact and the unavailability of the gold standards from \mwb{}, we further support the efficacy of our approach with a direct path-level comparison in the next section.}. 

Coverage is 100\% for all the baselines and the classifiers-based approaches. This is because at least one path is discovered for each node in the induction phase, thus resulting in at least one (possibly incorrect) hypernym for each node in the final taxonomy. This also serves to demonstrate that the initial projected taxonomy (cf. Section~\ref{sec:proj}) is reachable from every node in the target WCN.

In general, character-level models outperform their word-level counterparts. Char TFIDF significantly outperforms Word TFIDF for both entities and categories across all languages. Similarly, Char CNN outperforms Word CNN. Char LSTM outperforms Word LSTM for categories, but performs slightly worse for entities. We hypothesize that this is due to the difficulty in training character LSTM models over larger training sets. Entity training sets are much larger, as the number of Entity$\rightarrow$Category edges are significantly higher than the number of Category$\rightarrow$Category edges (usually by a factor of 10).
\vspace{-0.1cm}

\paragraph{Neural Models vs. TFIDF.}CNN-based models perform slightly better on average, followed closely by LSTM and TFIDF respectively. However, the training time for neural networks-based models is significantly higher than TFIDF models. For example, it takes approximately 25 hours to train the Char CNN model for French entities using a dedicated GPU. In contrast, the Char TFIDF model for the same data is trained in less than 5 minutes. Therefore, for the sake of efficiency, as well as to ensure simplicity and reproducibility across all languages, we choose Char TFIDF taxonomies as our final taxonomies for the rest of the evaluations. However, it is important to note that more accurate taxonomies can be induced by using our approach with neural-based models, especially if the accuracy of taxonomies is critical for the application at hand.

\begin{table}[t]
\small
\centering
\resizebox{\linewidth}{!}{
\begin{tabular}{l}
\toprule
\mwb{} \\
\midrule
\textbf{Patrimoine mondial en Équateur} $\leadsto$ Conservation de la nature $\rightarrow$ Écologie \\$\rightarrow$ Biologie $\rightarrow$ Sciences naturelles $\rightarrow$ Subdivisions par discipline \\$\rightarrow$ Sciences $\rightarrow$ Discipline académique \\$\rightarrow$ Académie $\rightarrow$ Concept philosophique \\
\midrule
\\ 
\toprule
Char TFIDF \\
\midrule
\textbf{Patrimoine mondial en Équateur} $\rightarrow$ \textbf{Patrimoine mondial en Amérique} \\$\rightarrow$ \textbf{Patrimoine mondial par continent} $\rightarrow$ \textbf{Patrimoine mondial} \\$\rightarrow$ \textbf{Infrastructure touristique} $\rightarrow$ \textbf{Lieu} $\leadsto$ Géographie \\$\rightarrow$ Discipline des sciences humaines et sociales \\$\rightarrow$ Sciences humaines et sociales $\rightarrow$ Subdivisions par discipline \\
\bottomrule
\end{tabular}
}
\caption{Samples of generalization paths for French categories from \mwb{} and Char TFIDF taxonomies. Correct path prefix (CPP) for each path is shown in bold.}
\label{table:sample_path}
\end{table}

\subsection{Path-level Evaluation}
In this section, we compare Char TFIDF against \mwb{} using a variety of path-quality measures. Path-based evaluation of taxonomies was proposed by \citet{guptarevisiting}, who demonstrated that good edge-level precision may not directly translate to good path-level precision for taxonomies. They proposed the average length of \textit{correct path prefix (CPP)}, i.e., the maximal correct prefix of a generalization path, as an alternative measure of quality of a taxonomy. Intuitively, it aims to capture the average number of upward generalization hops that can be taken until the first wrong hypernym is encountered. Following this metric, we randomly sample paths originating from 25 entities and 25 categories from the taxonomies, and annotate the first wrong hypernym in the upward direction. In total, we annotated 600 paths across six different languages for Char TFIDF and \mwb{} taxonomies. Table ~\ref{table:sample_path} shows examples of these generalization paths along with their CPPs\footnote{Same starting entities and categories are used for all taxonomies per language.}.

We report the average length of CPP (ACPP), as well as the average ratio of length of CPP to the full path (ARCPP). As an example, given the generalization path \textit{\textbf{apple}}$\rightarrow$\textit{\textbf{fruit}}$\leadsto$\textit{farmer}$\rightarrow$\textit{human}$\rightarrow$\textit{animal} with the \notisa{} edge \textit{fruit}$\leadsto$\textit{farmer}, the path length is 5, length of CPP is 2, and ratio of length of CPP to total path is 0.4 (i.e., $\frac{2}{5}$). 

Table~\ref{table:patheval} shows the comparative results. 
Char TFIDF taxonomies significantly outperform \mwb{} taxonomies, achieving higher average CPP lengths (ACPP) as well as higher average ratio of CPP to path lengths (ARCPP). Therefore, compared to the state-of-the-art \mwb{} taxonomies, Char TFIDF taxonomies are a significantly better source of generalization paths for both entities and categories across multiple languages.
\label{sec:patheval}

\begin{table}[t]
\centering

\resizebox{\linewidth}{!}{
\begin{tabular}{crccc|ccc}
\toprule
&  & \multicolumn{3}{c|}{Entity} & \multicolumn{3}{c}{Category} \\
Language & Method & AL & ACPP & ARCPP & AL & ACPP & ARCPP\\
\midrule
\multirow{2}{*}{French} 
& MultiWiBi & 8.24 & 2.96 & 0.49 & 8.92 & 3.6 & \textbf{0.56} \\
& Char TFIDF & 11.08 & \textbf{5.08} & 0.49 & 8.36 & \textbf{3.76} & 0.49\\
\midrule
\multirow{2}{*}{Italian} 
& MultiWiBi  & 7.36 & 2.68 & 0.45 & 14.84 & 3.72 & 0.27\\
& Char TFIDF  & 8.32 & \textbf{4.88} & \textbf{0.61} & 8.32 & \textbf{4.52} & \textbf{0.57}\\
\midrule
\multirow{2}{*}{Spanish} 
& MultiWiBi  & 7.04 & 3.08 & \textbf{0.55} & 12.08 & 4.08 & 0.36\\
& Char TFIDF & 12.8 & \textbf{5.0} & 0.48 & 12.76 & \textbf{5.28} & \textbf{0.48}\\
\midrule
\multirow{2}{*}{Arabic} 
& MultiWiBi  & 8.96 & 2.12& 0.31 & 14.64 & 4.12 & 0.31\\
& Char TFIDF  & 7.48& \textbf{5.88} & \textbf{0.81} & 6.96&\textbf{5.04} & \textbf{0.74}\\
\midrule
\multirow{2}{*}{Hindi} 
& MultiWiBi  & 7.72 & 1.88 & 0.27 & 7.4& 1.8 & 0.36\\
& Char TFIDF  & 10.28 & \textbf{4.92} & \textbf{0.47} & 8.0 & \textbf{2.44} & \textbf{0.38}\\
\midrule
\multirow{2}{*}{Chinese} 
& MultiWiBi  &7.4 & 2.56&0.47 & 8.0& 4.43&0.63\\
& Char TFIDF &6.32 &\textbf{3.92} &\textbf{0.68} &6.95 &\textbf{4.48} &\textbf{0.68}\\
\bottomrule
\end{tabular}
}
\caption{Comparison of average path length (AL), average length of correct path prefix (ACPP), and average ratio of CPP to path lengths (ARCPP).}
\label{table:patheval}
\end{table}

\begin{figure}[t]
\centering
\includegraphics[width=3.3in]{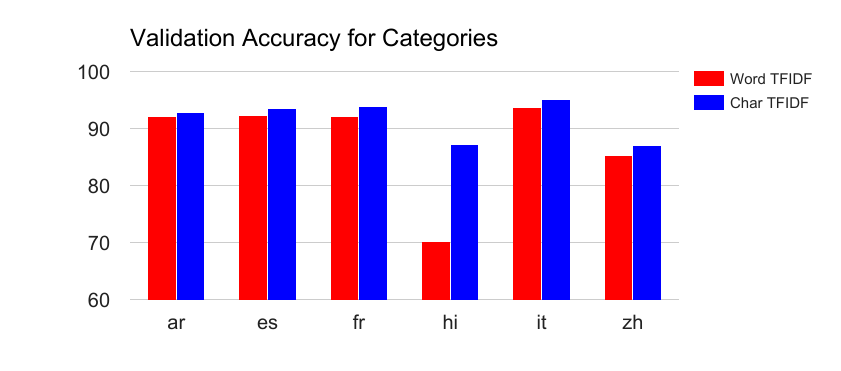}
\includegraphics[width=3.3in]{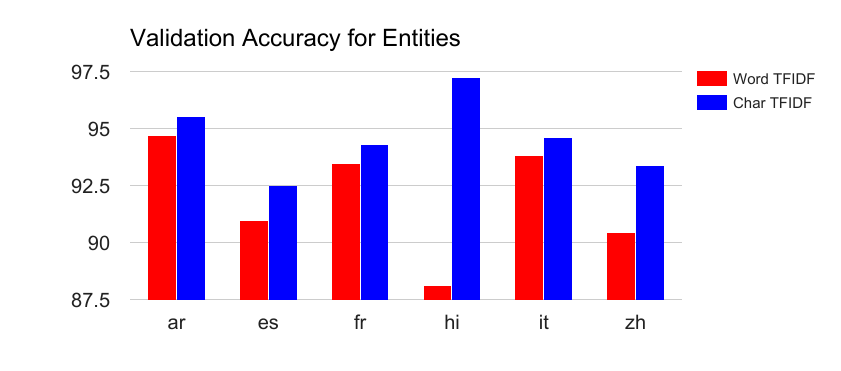}
\caption{Validation accuracies for Word TFIDF vs. Char TFIDF models.}
\label{fig:cat_test}
\end{figure}

\section{Analysis}
In this section, we perform additional analyses to gain further insights into our approach. More specifically, in Section~\ref{sec:wordvschar} and~\ref{sec:fpfn}, we perform an in-depth comparison of the Word TFIDF and Char TFIDF models. In section~\ref{sec:branch}, we show the effect of the parameter $k$, i.e., the number of paths discovered during optimal path search (cf. Induction Phase in Section~\ref{sec:indp}), on the branching factor and the precision of the induced taxonomies.

\subsection{Word vs. Character Models}
\label{sec:wordvschar}
To compare word and character-level models, we first report the validation accuracies for Word TFIDF and Char TFIDF models in Figure~\ref{fig:cat_test}, as obtained during the training phase\footnote{Validation set is constructed by randomly selecting 25\% of the edges with each label (i.e., \isa{} and \notisa{}) as discovered during the projection phase.} (cf. Section~\ref{sec:trainp}). Char TFIDF models significantly outperform Word TFIDF models, achieving higher validation accuracies across six different languages. The improvements are usually higher for languages with non-Latin scripts. This can be partly attributed to the error-prone nature of whitespace-based tokenization for such languages. For example, the word tokenizer for Hindi splits words at many accented characters in addition to word boundaries, thus leading to erroneous features and poor performance. In contrast, character-level models are better equipped to handle languages with arbitrary scripts, because they do not need to perform text tokenization.

\subsection{False Positives vs. False Negatives}
\label{sec:fpfn}
\begin{figure}[t!]
\centering	
\begin{subfigure}{.6\linewidth}
  \centering
\includegraphics[width=0.99\linewidth]{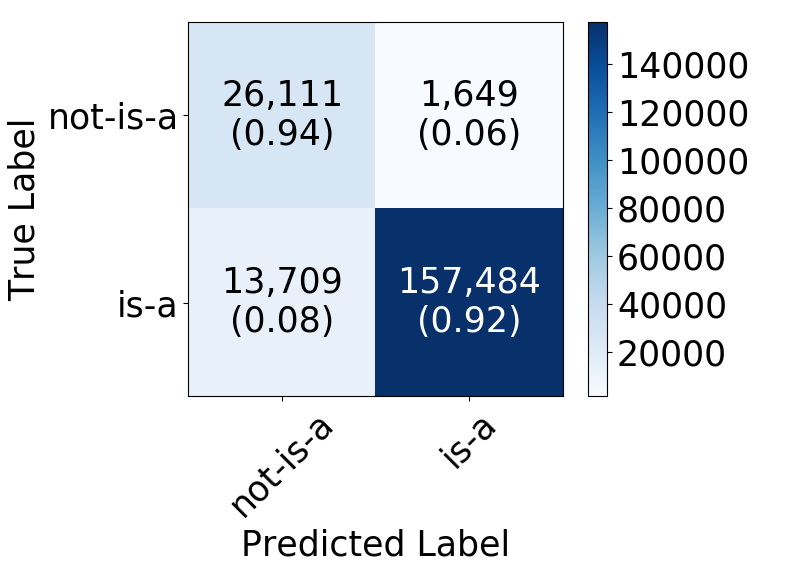}
\caption{\textbf{Word TFIDF}}
\end{subfigure}
\begin{subfigure}{.6\linewidth}
  \centering
\includegraphics[width=0.99\linewidth]{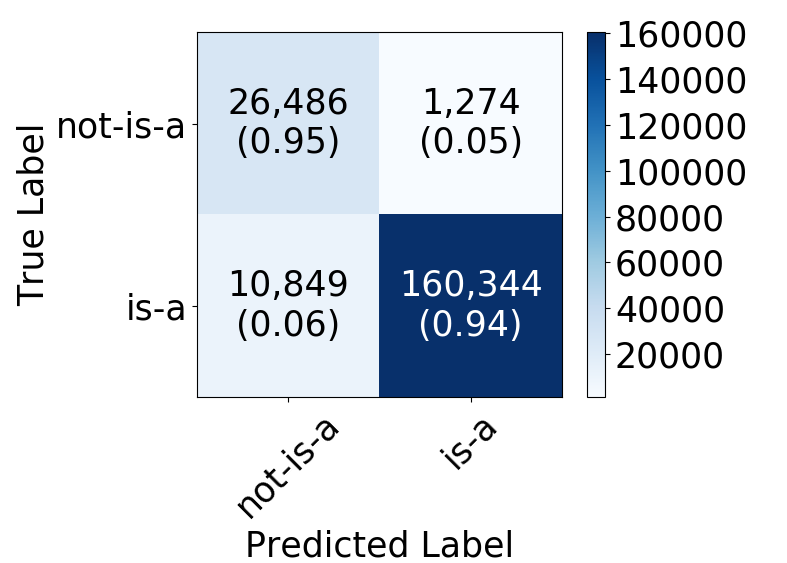}
\caption{\textbf{Char TFIDF}}
\end{subfigure}
\caption{Confusion matrices for Word TFIDF vs. Char TFIDF for French categories. Each cell shows the total number of edges along with the ratios in brackets.}
\label{fig:cm}
\end{figure}

To further compare word and character models, we focus on the specific case of French categories. In Figure~\ref{fig:cm}, we show the confusion matrices of Word TFIDF and Char TFIDF model computed using the validation set for French categories. While, in general, both models perform well, Char TFIDF outperforms Word TFIDF, producing fewer false positives as well as false negatives. We noticed similar patterns across most languages for both entities and categories.

We hypothesize that the superior performance of Char TFIDF is because character $n$-gram features incorporate the morphological properties computed at the sub-word level as well as word boundaries, which are ignored by the word-based features. To demonstrate this, we show in Tables~\ref{tab:fn} and~\ref{tab:fp} the top Word TFIDF and Char TFIDF features of a \notisa{} and an \isa{} edge. These edges are misclassified by Word TFIDF, but correctly classified by Char TFIDF. 

While Word TFIDF features are restricted to individual words, Char TFIDF features can capture patterns across word boundaries. For example the 6-gram feature ``\textit{ur spor}'' occurs in multiple hypernyms with different words: e.g., \textit{Commentateur sportif américain}, \textit{Entraîneur sportif américain} and \textit{Entraîneur sportif russe}. Such features incorporate morphological information such as plurality and affixes, which can be important for the detection of an \isa{} relationship. This is also evidenced by previous approaches that utilize multiple hand-crafted features based on such morphological information~\cite{suchanek2007yago,guptarevisiting}. Therefore, character-level models equipped with such features perform better at the task of WCN edge classification than their word-level counterparts. 

\begin{table}[t]
\centering

\resizebox{0.8\linewidth}{!}{
\begin{tabular}{c|c}
\toprule
Word TFIDF & Char TFIDF \\
\midrule
\textit{dolphins, dolphins, miami} & \textit{s dol, s dolp, es dol} \\ 
\textit{miami, entraîneur, des}  &  \textit{hins, dolph, hins d} \\
\end{tabular}
}
\caption{Top features for \notisa{} edge \textit{Entraîneur des Dolphins de Miami}$\leadsto$\textit{Dolphins de Miami}.}
\label{tab:fn}
\vspace{0.2cm} 
\resizebox{0.85\linewidth}{!}{
\begin{tabular}{c|c}
\toprule
Word TFIDF & Char TFIDF \\
\midrule
\textit{dolphins, américain, miami} &  \textit{ur spor, r sport, eur sp}\\ 
\textit{entraîneur, sportif, entraîneur}  & \textit{tif am, if am, if amé} \\
\end{tabular}
}
\caption{Top features for \isa{} edge \textit{Entraîneur des Dolphins de Miami}$\rightarrow$\textit{Entraîneur sportif américain}.}
\label{tab:fp}
\end{table}

\subsection{Precision vs. Branching Factor}
\label{sec:branch}

Along with standard precision/recall measures, structural evaluation also plays an important role in assessing the quality of a taxonomy. One of the important structural properties of a taxonomy is the \textit{branching factor}, which is defined as the average out-degree of the nodes in the taxonomy. Taxonomies with higher branching factors are desirable, because they are better equipped to account for multiple facets of a concept or an entity (e.g., \textit{Bill Gates} is both a philanthropist and an entrepreneur). 

However, there is usually a trade-off between branching factor and precision in automatically induced taxonomies~\cite{velardi2013ontolearn}. Higher branching factor typically results in lowering of precision due to erroneous edges with lower scores being added to the taxonomy. Prioritizing the precision over the branching factor or vice-versa is usually determined by the specific use case at hand. Therefore, it is desirable for a taxonomy induction method to provide a control mechanism over this trade-off.

\begin{figure}
\centering
\includegraphics[width=2.4in]{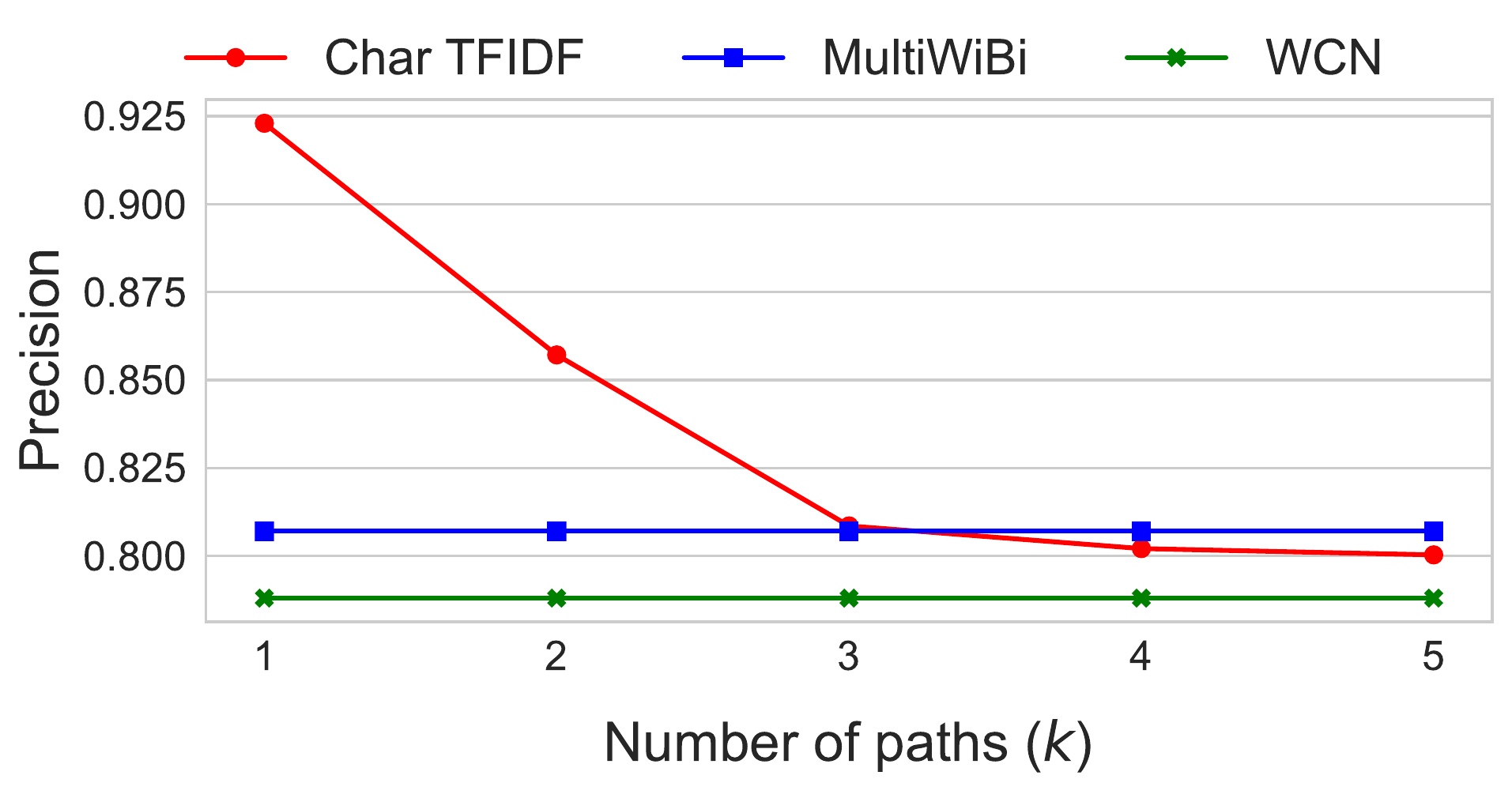}
\includegraphics[width=2.4in]{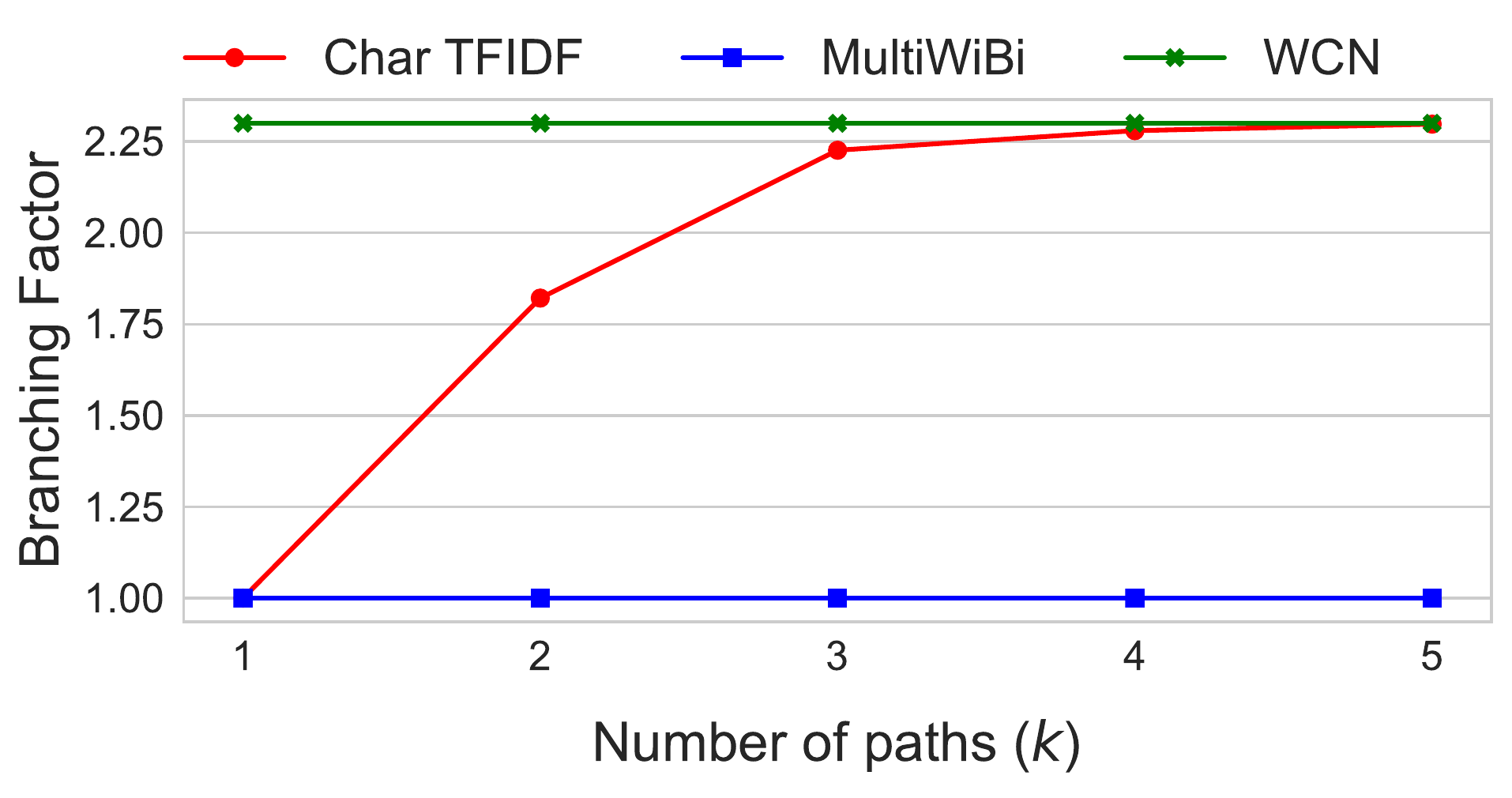}

\caption{Precision vs. branching factor for different number of paths ($k$) in the Induction phase (cf. Section~\ref{sec:indp}).}
\label{fig:bf}
\end{figure}

In our approach, the number of paths discovered ($k$) in the optimal path search (cf. Section~\ref{sec:indp}) serves as the parameter for controlling this trade-off. As $k$ increases, the branching factor of the induced taxonomy increases because more paths per term are discovered. To demonstrate this effect, we plot the values of precision and branching factor of Char TFIDF taxonomies for varying values of $k$ for French categories\footnote{Similar effects are observed for both entities and categories for all languages.} in Figure~\ref{fig:bf}. Precision and branching factors for \mwb{} taxonomies and the original WCN are also shown for comparison purposes. 

Char TFIDF significantly outperforms \mwb{}, either achieving higher precision ($k$$\le$$2$) or higher branching factor ($k$$\ge$$2$). At $k$$=$$2$, Char TFIDF presents a sweet spot, outperforming \mwb{} in both precision and branching factor. For $k$$\ge$$3$, Char TFIDF taxonomies start to resemble the original WCN, because most of the WCN edges are selected by optimal path discovery. This experiment demonstrates that in contrast to \mwb{}'s fixed set of heuristics, our approach provides a better control over the branching factor of the induced taxonomies.

\section{Related Work and Discussion}
\label{sec:wibivschar}
The large-scale and high quality of Wikipedia content has enabled multiple approaches towards knowledge acquisition and taxonomy induction over the past decade. Earlier attempts at taxonomy induction from Wikipedia focus on the English language. WikiTaxonomy, one of the first attempts to taxonomize Wikipedia, labels English WCN edges as \isa{} or \notisa{} using a cascade of heuristics based on hand-crafted features~\cite{ponzetto2008wikitaxonomy}. WikiNet extends WikiTaxonomy by expanding \notisa{} relations into more fine-grained relations such as meronymy (i.e., \textit{part-of}) and geo-location (i.e., \textit{located-in}). YAGO induces a taxonomy by linking Wikipedia categories to WordNet synsets using a set of simple heuristics~\cite{suchanek2007yago,hoffart2013yago2}. DBPedia provides a fully-structured knowledge representation for the semi-structured content of Wikipedia, which is further linked to existing knowledge bases such as YAGO and OpenCyc~\cite{auer2007dbpedia,dbpedia2}. More recently,~\citet{guptarevisiting} induce a unified taxonomy of entities and categories from English WCN using a novel set of high-precision heuristics that classify WCN edges into \isa{} and \notisa{}.   

A second line of work aims to extend the taxonomy induction process to other languages by exploiting the multilingual nature of Wikipedia content. MENTA, a large-scale multilingual knowledge base, is induced by linking WordNet with WCN of different languages into a unified taxonomy~\cite{de2010menta}. The most recent and the most notable effort towards this direction is \mwb{}~\cite{flati2016multiwibi}.  \mwb{} first simultaneously induces two separate taxonomies for English, one for pages and one for categories. To this end, it exploits the idea that information contained in pages are useful for taxonomy induction over categories and vice-versa. To induce taxonomies for other languages, \mwb{} employs a set of complex heuristics, which utilize hand-crafted features (such as textual and network topology features) and a probabilistic translation table constructed using the interlanguage links. 

Our approach borrows inspiration from many of the aforementioned approaches. First, similar to WikiTaxonomy and ~\citet{guptarevisiting}, our approach also classifies WCN edges into \isa{} or \notisa{}. Second, similar to \mwb{}, our approach also projects an English taxonomy into other languages using the interlanguage links. However, unlike these approaches, our approach does not employ any heuristics or hand-crafted features. Instead, it uses text classifiers trained on an automatically constructed dataset to assign edge weights to WCN edges. Taxonomic edges are discovered by running optimal path search over the WCN in a fully-automated and language-independent fashion.   

Our experiments show that taxonomies derived using our approach significantly outperform the state-of-the-art taxonomies, derived by \mwb{} using more complex heuristics. We hypothesize that it is because our model primarily uses categories as hypernyms, whereas \mwb{} first discovers hypernym lemmas for entities using potentially noisy textual features derived from unstructured text. Categories have redundant patterns, which can be effectively exploited using simpler models. This has also been shown by~\citet{guptarevisiting}, who use simple high-precision heuristics based on the lexical head of categories to achieve significant improvements over \mwb{} for English.

Additionally, for taxonomy induction in other languages, \mwb{} uses a probabilistic translation table, which is likely to introduce further noise. The high-precision heuristics of~\citet{guptarevisiting} are not easily extensible to languages other than English, due to the requirement of a syntactic parser for lexical head detection. In contrast, our approach learns such features from automatically generated training data, hence resulting in high-precision, high-coverage taxonomies for all Wikipedia languages.

\section{Conclusion}
\label{sec:conc}
In this paper, we presented a novel approach towards multilingual taxonomy induction from Wikipedia. Unlike previous approaches which are complex and heuristic-heavy, our approach is simpler, principled and easy to replicate. Taxonomies induced using our approach outperform the state of the art on both edge-level and path-level metrics across multiple languages. Our approach also provides a parameter for controlling the trade-off between precision and branching factor of the induced taxonomies. A key outcome of this work is the release of our taxonomies across 280 languages, which are significantly more accurate than the state of the art and provide higher coverage.

\bibliographystyle{aaai}
\bibliography{mybib}

\end{document}